\documentclass{article}
\usepackage{spconf,amsmath,graphicx,hyperref}
\usepackage{booktabs,multirow,pifont,makecell}

\def\L{{\cal L}}

\newcommand{\shortname}{SAGE-LD}

\title{SAGE-LD: Towards Scalable and Generalizable End-to-End Language Diarization via Simulated Data Augmentation}
\name{Sangmin Lee, Woongjib Choi, Jihyun Kim, Hong-Goo Kang}
\address{Dept. of Electrical \& Electronic Engineering, Yonsei University, Seoul, South Korea}
\begin{document}
%
\maketitle

\begin{abstract}
In this paper, we present a neural spoken language diarization model that supports an unconstrained span of languages within a single framework. Our approach integrates a learnable query-based architecture grounded in multilingual awareness, with large-scale pretraining on simulated code-switching data. By jointly leveraging these two components, our method overcomes the limitations of conventional approaches in data scarcity and architecture optimization, and generalizes effectively to real-world multilingual settings across diverse environments.
Experimental results demonstrate that our approach achieves state-of-the-art performance on several language diarization benchmarks, with a relative performance improvement of 23\% to 52\% over previous methods. We believe that this work not only advances research in language diarization but also establishes a foundational framework for code-switching speech technologies.
\end{abstract}

\vspace{-7pt}
\section{Introduction}

Language diarization (LD) refers to the task of determining \textit{which language is spoken at a given point in time} within an audio stream. This task is particularly important in code-switching (CS) scenarios, where a single speaker alternates between languages within or across utterances. 
Such behavior introduces significant challenges for multilingual speech processing systems, as phonetic, syntactic, and lexical properties can differ significantly across languages.
In this context, accurate LD enables the decomposition of CS utterances into monolingual segments, thereby enabling the application of language-specific downstream systems that generally outperform multilingual models in constrained conditions.

Previous research on LD has primarily progressed along two directions. 
The first approach integrates LD as a subcomponent of code-switching automatic speech recognition (CS-ASR)~\cite{csasr_ld1, csasr_ld2}, where language boundaries are either implicitly modeled or explicitly annotated to support multilingual transcription. Although methods using this strategy can achieve high diarization accuracy, they assume a single language pair (e.g., Mandarin–English) environment, which limits their applicability in broader multilingual or general contexts.
The second line of work frames LD as a standalone task drawing parallels to speaker diarization, a task that aims to segment speech by speaker identity. These types of LD systems typically adopt multi-stage pipelines~\cite{displace, ld_cluster2} consisting of data processing, feature extraction, followed by clustering, or leverage end-to-end neural diarization methods~\cite{eeld1, eeld2}, enabling a single model to process multiple languages.

Recently, efforts have focused on developing general-purpose LD models that handle multiple languages within a single framework. The DISPLACE challenge~\cite{displace} introduced a benchmark for Indic–English LD in conversational scenarios, marking a milestone toward broader LD modeling. However, performance under this setup still lags significantly behind that in speaker diarization. Complementary work has been done on Bantu–English LD~\cite{eeld2} using a broadcast corpus, but performance lags behind the DISPLACE benchmark, and coverage was restricted to a fixed set of languages.

To address these limitations, we propose \shortname,\footnote{\textbf{S}calable \textbf{A}nd \textbf{G}eneralizable \textbf{E}nd-to-end \textbf{L}anguage \textbf{D}iarization}\footnote{Github: \url{https://github.com/sanghyang00/sage-ld}} a comprehensive framework for end-to-end language diarization that supports an unbounded number of languages.
Inspired by instance segmentation on various domains~\cite{mask2former,eend_m2f} and other multilingual speech technologies~\cite{langgroup1,langgroup2}, we combine multilingual acoustic features, a contextual encoder, and a decoder with learnable language queries. 
Then, we construct a simulated corpus exceeding 100 hours of speech across more than 20 language pairs to pretrain the model, providing generalized diarization capabilities for detecting language shifts in diverse matrix–embedded language configurations. Finally, the model is adapted to a small amount of annotated real-world data to capture domain-specific characteristics.

In experiments, \shortname~achieves state-of-the-art performance across several LD benchmarks. Notably, our method consistently shows superior results in both long-form conversational and short-form broadcast settings, with relative improvements of 30\% and 52\%, respectively. These results demonstrate the robustness and versatility of our approach across a variety of language and acoustic conditions. We anticipate our work will pave the way for broader advancements in language diarization, especially in language coverage, and facilitate improved integration with massively multilingual code-switching speech technologies.
\section{Related Work}
\vspace{-7pt}
\label{sec:rw}
There are broadly two approaches for LD: multi-stage and end-to-end systems.
Multi-stage LD splits the task into preprocessing, feature extraction, clustering, and postprocessing~\cite{displace,ld_cluster2}. These systems typically follow a modular pipeline in which each component is independently designed for optimal performance within its scope. However, their reliance on a fixed-length sliding window for feature extraction makes them better suited for long-form inputs. 
In contrast, end-to-end LD uses a single model to segment speech by language. Recent approaches~\cite{eeld1,eeld2} utilize speech self-supervised models (S3Ms) with segmentation heads for predicting language labels over time, treating LD as a frame-level multiclass classification problem operating on contextual S3M features. However, these methods assume a fixed language set, limiting generalizability to larger or open-ended language inventories.

Beyond the model design, progress in LD research has also been constrained by data availability. Existing datasets for the task include SEAME~\cite{seame}, MSCS~\cite{mscs}, MERLIon CCS~\cite{merlion}, the South African (SA) Soap Opera corpus~\cite{southafrican}, and the DISPLACE challenge corpus~\cite{displace}. However, a major limitation is that large-scale corpora tend to focus on single language pairs (primarily Mandarin-English), while others covering more language pairs remain relatively small in size. This scarcity of multilingual LD data makes it challenging to develop models that generalize across diverse languages, highlighting the need for novel methods that can handle unbounded languages and diverse environments.

\vspace{-10pt}
\section{Proposed Method}
\vspace{-7pt}
As highlighted in Section~\ref{sec:rw}, a key challenge in LD is the \textbf{\textit{lack of generalizability}} across languages and conditions. We address this issue through: (1) architectural refinement to maximize the flexibility of the model, and (2) simulated data augmentation to relieve the data scarcity problem, thereby providing a strong foundation for real-world applications.

\vspace{-10pt}
\subsection{End-to-End Language Diarization Model}
\vspace{-5pt}
\begin{figure}[t!]
    \centering
    \includegraphics[width=0.35\textwidth]{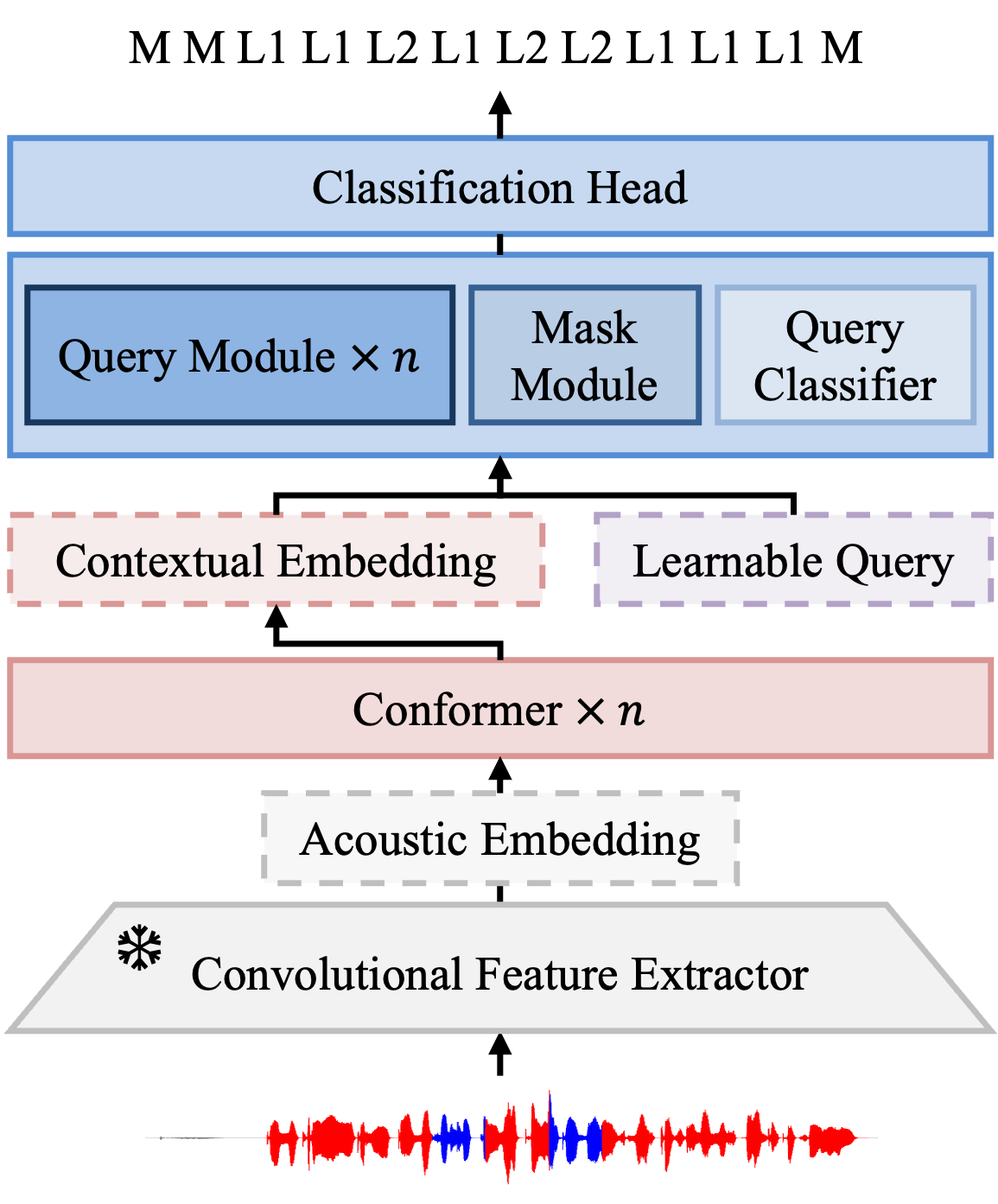}
    \vspace{-7pt}
    \caption{Architecture of the SAGE-LD, and we set $n=6$. The model comprises three modules: feature extractor, contextual encoder, and decoder with learnable language queries.}
    \label{fig:model}
    \vspace{-10pt}
\end{figure}

\shortname~processes raw waveforms and produces diarization outputs through three components: a feature extractor, a contextual encoder, and a masked attention decoder with learnable queries, as illustrated in Fig.~\ref{fig:model}. We describe the design choices and the rationale for each component in the following.

\vspace{2pt}
\noindent\textbf{Multilingual Feature Extractor.}
Multilingual S3Ms~\cite{xlsr,mms} have demonstrated strong cross-lingual generalizability, providing robust features from raw waveforms across diverse environments and languages. They first extract frame-level \textbf{\textit{acoustic features}} through a convolutional feature extractor (e.g., every 25 ms), which are then refined into \textbf{\textit{contextual representations}} by Transformer layers~\cite{transformer,s3m_layer}. 
However, existing S3Ms are primarily trained on monolingual (non-CS) utterances, which causes their contextual embeddings from CS utterances to mix linguistic information after the Transformer layers. Thus, directly leveraging these features for LD might be sub-optimal. To circumvent this, our feature extractor module only utilizes the convolutional layers of the S3M, deliberately omitting the Transformer layers. This approach extracts language-agnostic acoustic features, capturing more universal characteristics of speech. Specifically, we leverage the pretrained feature extractor module of MMS~\cite{mms}.

\vspace{2pt}
\noindent\textbf{Contextual Encoder.}
\label{sec:cenc}
To capture language-aware context, we stack Conformer~\cite{conformer} layers.
Unlike most previous LD models that aggregate acoustic features using sliding windows~\cite{displace,ld_cluster2} or feature pooling~\cite{eeld1} to coarsen features beyond 25 ms, we avoid such aggregation. The rationale here is that Conformer layers directly model frame interactions through convolutional modules, which effectively expand the model's receptive field while serving as an implicit aggregation mechanism.
As a result, adding an extra pooling step offers little computational benefit while discarding temporal cues crucial for LD performance.
Our design further leverages the large-scale pretraining described in Section~\ref{sec:pt}, enabling our model to directly learn fine-grained acoustic features and refine them into contextual embeddings for robust LD.

\begin{figure}[t!]
    \centering
    \includegraphics[width=0.4\textwidth]{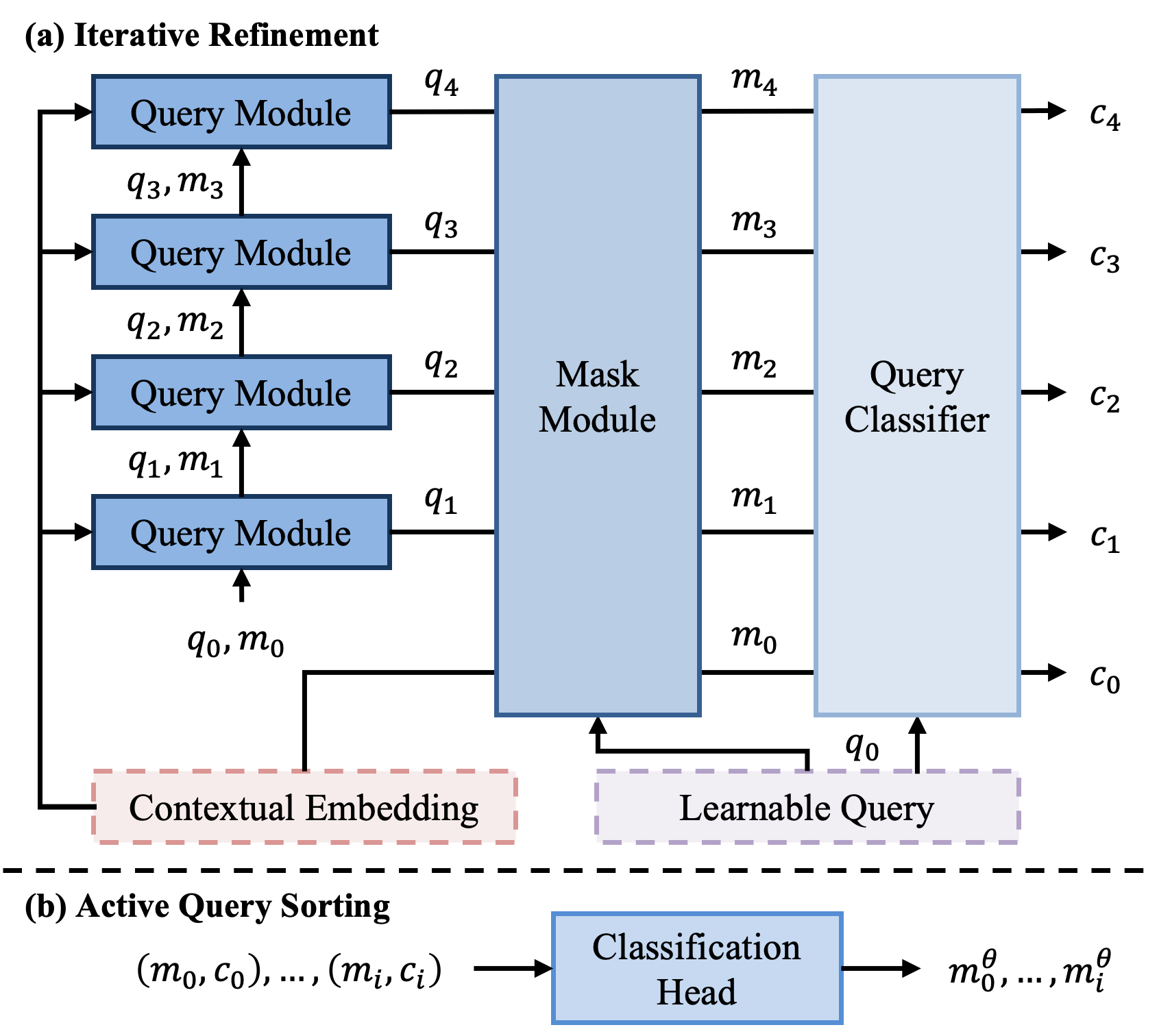}
    \vspace{-7pt}
    \caption{Architecture of the decoder with $n=4$. Each query $q_i$, mask $m_i$, and activity $c_i$ is iteratively refined, and the classification head sorts active queries to generate a prediction $m_i^\theta$.}
    \label{fig:decoder}
    \vspace{-10pt}
\end{figure}

\vspace{2pt}
\noindent\textbf{Masked Attention Decoder.}
We adopt a masked attention decoder with learnable queries, a component widely used in segmentation models~\cite{mask2former, eend_m2f}. It consists of Transformer decoder-based multiple query modules, a mask module built from three feedforward layers, and a query classifier of a single linear layer. We further introduce two task-specific modifications. First, we use a small number of queries (five), since LD typically involves CS between only a few languages, making larger query sets unnecessary. Second, we frame LD as a multiclass classification problem, where one query slot is explicitly reserved for voice activity detection (VAD).
The decoding process proceeds as follows. First, the mask module combines the initial query $q_0$ with contextual embeddings to predict a mask $m_0$ and query activity $c_0$. Then the query $q_i$, mask $m_i$, and query activity $c_i$ are iteratively refined, aggregating contextual information. In each step, the classification head sorts the active queries based on the query activity $c_i$, and language prediction $m_i^\theta$ is computed from the mask $m_i$. The overall procedure is depicted in Fig.~\ref{fig:decoder}.

\vspace{-10pt}
\subsection{Data Augmentation with Simulated Utterances}
\vspace{-3pt}
\label{sec:unicom}
Then we aim to leverage large-scale training data to maximize the performance of \shortname. Inspired by speaker diarization pretraining, we hypothesize that LD pretraining on simulated utterances can be beneficial. 
However, a critical challenge arises: disentangling speaker shifts from language shifts. Naively concatenating monolingual utterances can superficially mimic CS but conflates language boundaries with speaker changes, potentially leading the model to perform speaker diarization rather than LD.
To address this, we simulate CS utterances using UniCoM~\cite{unicom}, since it utilizes a voice conversion (VC) model~\cite{knnvc} to unify the speaker identity over utterances. This approach effectively decouples language and speaker transitions, yielding supervised data for pretraining.

\vspace{-10pt}
\subsection{Language-Aware Training Strategy}
\vspace{-3pt}
\label{sec:pt}
Subsequently, drawing on insights from cross-lingual transfer, we adopt a two-stage training strategy that consists of language family-based pretraining and dataset-specific adaptation. 
In the first stage, we construct a large-scale simulated corpus covering various unique language pairs within a target language family, using the method explained in Section~\ref{sec:unicom}.
This pretraining equips the model with general diarization capabilities and transfers language knowledge into the encoder, enabling robust detection of language shifts across diverse matrix–embedded configurations. Moreover, grouping languages by family improves knowledge transfer since related languages share linguistic features, a strategy which has been shown to yield strong empirical gains in multilingualism research~\cite{langgroup1,langgroup2}.
In the second stage, the model is adapted to a small set of real-world LD data, allowing it to capture domain-specific characteristics while leveraging the generalized diarization capabilities acquired during pretraining.

\begin{table*}[!t]
\centering
\caption{LD performance comparison across models. DER values are reported with their breakdown (False Alarm / Miss / Confusion) inside parentheses. * denotes closed-source; results are from the original paper. For multi-stage models, only the feature extractor size is reported, with extra parameters indicated by a + symbol, as some subcomponent details are unavailable.}
\vspace{2pt}
\resizebox{0.95\textwidth}{!}{%
\begin{tabular}{l|c|c|c|cl|c|cl|c|cl} \toprule
\multirow{2}{*}{Model} & \multirow{2}{*}{E2E} & \multirow{2}{*}{Size} 
& \multicolumn{3}{c|}{DISPLACE-D} 
& \multicolumn{3}{c|}{DISPLACE-E} 
& \multicolumn{3}{c}{SA Soap Opera} \\ 
 &  &  & Ideal & \multicolumn{2}{c|}{Practical} & Ideal & \multicolumn{2}{c|}{Practical} & Ideal & \multicolumn{2}{c}{Practical} \\ \midrule\midrule

DISPLACE 2024~\cite{displace} & X & 74M+ 
& 33.20 & 38.01 & {\scriptsize (4.66/3.99/29.36)} 
& 23.14 & 28.46 & {\scriptsize (2.64/5.15/20.66)} 
& N/A & \multicolumn{2}{c}{N/A} \\

TalTech-IRIT-LIS\textsuperscript{*}~\cite{ld_cluster2} & X & 600M+ 
& - & 28.20 & {\scriptsize (-/-/-)} 
& - & 27.60 & {\scriptsize (-/-/-)} 
& N/A & \multicolumn{2}{c}{N/A} \\

Mishra et al.~\cite{eeld1} & O & 108M 
& 27.04 & 29.24 & {\scriptsize (4.83/3.30/21.11)} 
& 25.80 & 28.14 & {\scriptsize (4.85/3.63/19.65)} 
& 65.44 & 65.53 & {\scriptsize (0.16/0.00/65.37)} \\

Frost et al.~\cite{eeld2} & O & 315M 
& 17.11 & 27.98 & {\scriptsize (7.75/3.74/16.49)} 
& 18.60 & 29.46 & {\scriptsize (6.81/5.33/17.32)} 
& 35.53 & 35.30 & {\scriptsize (6.98/2.77/25.54)} \\ \midrule

SAGE-LD (w/o PT) & O & \multirow{2}{*}{72M} 
& \underline{16.90} & \underline{22.82} & {\scriptsize (3.54/3.58/15.70)} 
& \underline{16.00} & \underline{23.24} & {\scriptsize (3.16/3.09/14.99)} 
& \underline{14.49} & \underline{18.55} & {\scriptsize (2.37/3.00/13.18)} \\

\textbf{SAGE-LD (w/ PT)} & O & 
& \textbf{15.18} & \textbf{21.37} & {\scriptsize (3.03/3.69/14.64)} 
& \textbf{14.63} & \textbf{18.03} & {\scriptsize (2.28/1.80/13.95)} 
& \textbf{13.05} & \textbf{16.92} & {\scriptsize (2.81/2.03/12.07)} \\ \bottomrule
\end{tabular}
}
\label{performance_comp}
\vspace{-10pt}
\end{table*}

\vspace{-10pt}
\section{Experiments}
\vspace{-7pt}
\subsection{Training Criteria}
\vspace{-3pt}

We trained \shortname~using three losses: diarization loss ($\L_{dia}$), overlap loss ($\L_{ovr}$), and activation loss ($\L_{act}$). They are defined as follows, with all loss coefficients set to 1:
\vspace{-3pt}
\begin{equation}
    \L_{total} = \lambda_{dia}\L_{dia}+\lambda_{ovr}\L_{ovr}+\lambda_{act}\L_{act}.
    \label{total_loss}
    \vspace{-2pt}
\end{equation}
For the diarization loss, we adopt focal loss~\cite{focal}, a variant of binary cross-entropy (BCE) loss, which facilitates frame-level classification that focuses on hard examples. Here, $m$ denotes the ground truth label, and $m_i^\theta$ denotes the predicted label. The vector $\alpha_d$ assigns weights to VAD, matrix, and embedded languages, with a value of 3 for embedded languages and 1 for all others, and $\gamma_d = 0.25$. The loss is formulated as follows:
\vspace{-3pt}
\begin{multline}
\small
\L_{dia} = - \alpha_d \big(
    (1-m_i^\theta)^{\gamma_d} m \log(m_i^\theta) \; + \\
    {m_i^\theta}^{\gamma_d} (1-m)\log(1-m_i^\theta)
\big)
\vspace{-2pt}
\end{multline}
For the overlap loss, we adopt focal Tversky loss~\cite{focal_tversky}, a variant of dice loss~\cite{dice}, to handle imbalanced diarization due to the sparse occurrence of embedded languages. It complements the diarization loss by promoting greater overlap between predicted and ground truth labels while emphasizing accurate diarization of embedded languages. Specifically, $TP_{(m_i^\theta,m)}$, $FP_{(m_i^\theta,m)}$, and $FN_{(m_i^\theta,m)}$ denote true positives, false positives, and false negatives between the predicted label and the ground truth label. We set $\alpha_o = 0.7$ and $\beta = 0.3$ for embedded languages, $\alpha_o, \beta = 0.5$ for other classes, and $\gamma_o = 0.75$. The loss is formulated as follows:
\vspace{-3pt}
\begin{equation}
    \small
    \L_{ovr} = (1 - \frac{TP_{(m_i^\theta,m)}}{TP_{(m_i^\theta,m)} + \alpha_o FP_{(m_i^\theta,m)} + \beta FN_{(m_i^\theta,m)}})^{\gamma_o}.
\end{equation}
Finally, the activation loss distinguishes active and inactive queries. It is computed as BCE between the predicted and ground truth query activities $c$. Then, the loss is as follows:
\vspace{-3pt}
\begin{equation}
    \L_{act} = - (c \log c_i + (1 - c) \log (1 - c_i)).
    \vspace{-2pt}
\end{equation}
Additionally, following prior work~\cite{mask2former,eend_m2f}, we applied Hungarian matching to ensure permutation-invariant training among active queries, with the cost matrix formulated identically to Eq.~(\ref{total_loss}), and employed deep supervision during training.

\vspace{-10pt}
\subsection{Dataset Preparation}
\vspace{-3pt}
In pretraining, we simulated 100 hours each of Indic– and Bantu–English CS utterances from the FLEURS-R~\cite{fleursr} corpus, aligning the languages with each adaptation dataset. We further replaced the UniCoM's VC module with SeedVC~\cite{seedvc} to improve quality on non-European languages, as the original VC module was trained only on English.
Subsequently, simulated utterances were augmented with room impulse responses (RIRs) and background noise. Background noise was sampled from DEMAND~\cite{demand} and RIRs were drawn from the BUT database~\cite{but}. Each utterance had a 50\% probability of being augmented with both noise and RIR, with the signal-to-noise ratio randomly selected from 5, 10, 15, or 20 dB.

We used two public LD corpora to adapt and evaluate \shortname. The first, DISPLACE 2024~\cite{displace}, is a long-form conversational dataset with noisy environments, covering several Indic languages and English. As it is a dataset from a challenge, only the dev and test sets are available with different characteristics. Therefore, we treated them as distinct datasets: DISPLACE-D and DISPLACE-E. The second, SA Soap Opera~\cite{southafrican}, consists of short broadcast clips featuring four African languages with English. Each corpus was split into adaptation and evaluation sets in a 7:3 ratio.

\begin{table}[!t]
\vspace{-12pt}
\centering
\caption{Impact of feature pooling (or frame rate) in DER.}
\vspace{2pt}
\resizebox{0.9\columnwidth}{!}{%
\begin{tabular}{c|ccc} \toprule
Frame Rate & DISPLACE-D & DISPLACE-E & SA Soap Opera \\ \midrule\midrule
25 ms       &       \textbf{21.37}       &       \textbf{18.03}        &         \textbf{16.92}         \\
105 ms      &       21.97       &       18.95        &         17.65         \\
205 ms      &       21.91       &       19.09        &        18.25          \\ \bottomrule
\end{tabular}
}
\label{frame_rate_comp}
\vspace{-15pt}
\end{table}

\vspace{-12pt}
\subsection{Quantitative Evaluation}
\vspace{-5pt}
As shown in Table~\ref{performance_comp}, \shortname~achieves state-of-the-art results across all benchmarks, consistently outperforming prior works by a substantial margin. This performance is particularly notable given that our model uses the smallest number of parameters. Improvements are especially pronounced on the SA Soap Opera corpus, where short utterances with minimal contextual information pose significant challenges. Despite these difficulties, \shortname~surpasses previous LD methods, demonstrating the robustness of our approach. In the ideal scenario where VAD operates perfectly and the task focuses solely on discrimination between spoken languages, \shortname~still surpasses prior LD models. Moreover, simulated pretraining consistently improves performance over training from scratch, confirming the effectiveness of our approach.

Multi-stage models (`X' in the E2E column) rely on a fixed sliding window over 5 seconds, making them unsuitable for the SA Soap Opera dataset, where utterances are too short. In contrast, \shortname~performs well on short utterances while consistently outperforming two-stage models on long-form speech. On the other hand, end-to-end models (`O' in the E2E column) perform diarization at finer temporal resolutions using speech embeddings. In this case, prior work~\cite{eeld1} reported models only predicting a single language on short utterances, whereas \shortname~maintains robust performance.

\newcommand{\cmark}{\ding{51}}
\newcommand{\xmark}{\ding{55}}
\begin{table}[!t]
\vspace{-12pt}
\centering
\caption{Impact of loss design in DER.}
\vspace{2pt}
\resizebox{0.9\columnwidth}{!}{%
\begin{tabular}{cc|c|c|c} \toprule
\multicolumn{2}{c}{Loss} & \multicolumn{3}{c}{Dataset}                     \\
Focal      & Focal Tversky     & DISPLACE-D & DISPLACE-E & SA Soap Opera \\ \midrule\midrule
    \xmark       &      \xmark       &       23.77       &       19.67        &         18.38         \\
    \xmark       &      \cmark       &       23.34       &       18.63        &        17.31          \\
    \cmark       &      \xmark       &       22.46       &       18.21        &         18.17         \\
    \cmark       &      \cmark       &       \textbf{21.37}       &       \textbf{18.03}        &       \textbf{16.92}           \\ \bottomrule
\end{tabular}
}
\label{loss_comp}
\vspace{-15pt}
\end{table}
\vspace{-10pt}
\subsection{Ablation Study}
\vspace{-5pt}
\noindent\textbf{Impact of Feature Pooling.}
In Table~\ref{frame_rate_comp}, results show that applying additional attentive pooling over 25 ms acoustic features generally degrades performance. This finding aligns with the discussion in Section~\ref{sec:cenc} and supports our claim that pooling incurs information loss and excessively enlarges the receptive field, negatively affecting diarization quality.

\vspace{2pt}
\noindent\textbf{Impact of Loss Design.}
In Table~\ref{loss_comp}, we compare BCE against focal loss and dice against focal Tversky. Results indicate that replacing either loss individually yields notable gains, while adopting both together achieves the best results. This demonstrates the effectiveness of our LD-aware loss design, which accounts for the sparse occurrence of embedded languages.
\vspace{-10pt}
\section{Conclusion}
\vspace{-5pt}
In this paper, we present \shortname, a comprehensive framework for language diarization. Our approach consists of a generalizable end-to-end model capable of handling an unbounded number of languages, leveraging a multilingual feature extractor, contextual modeling, and iterative decoding with learnable queries. We further observe performance gains through a language-aware pretraining scheme using simulated code-switching utterances. Experimental results show that \shortname~achieves state-of-the-art performance across multiple language diarization benchmarks, regardless of recording environment and language. We believe \shortname~can advance research in language diarization and support broader developments in code-switching speech technology.

\vfill\pagebreak


\bibliographystyle{IEEEbib}
\bibliography{strings,refs}

\end{document}